%% file: acl_latex.tex
\documentclass[11pt]{article}

\usepackage[final]{acl}

\usepackage{times}
\usepackage{latexsym}

\usepackage[T1]{fontenc}

\usepackage[utf8]{inputenc}

\usepackage{microtype}
\usepackage{comment}
\usepackage{inconsolata}
\usepackage{comment}
\usepackage{graphicx}

\title{Lexical Tone and Word Order Matter in Translation:\\ Prosodic Correspondence Across Languages}

\author{
  \normalfont
  \textbf{Haopeng Xie} \quad
  \textbf{Ismail Rasim Ulgen} \quad
  \textbf{Shengying Zhao} \\
  \textbf{Berrak Sisman} \quad
  \textbf{Philipp Koehn} \\
  Johns Hopkins University \\
  \texttt{hxie25@jh.edu} \quad
  \texttt{\{iulgen1,szhao63,sisman,phi\}@jhu.edu}
}

\usepackage{booktabs}
\usepackage{float}
\usepackage{placeins}
\usepackage{graphicx}
\graphicspath{{figures/}}

\begin{document}
\maketitle
\begin{abstract}

Prosody conveys emphasis, emotion, and intent, yet its correspondence across translation remains underexplored, especially when languages differ in lexical tone and word order. To our knowledge, we present the first systematic word-level analysis of prosodic correspondence in translated speech spanning tonal and non-tonal Asian languages, with comparisons to European-language findings and explicit analyses of lexical tone and word reordering. Building on prior evidence of fine-grained prosody correspondence in European-language pairs, we investigate how these linguistic factors relate to word-level pitch and energy correspondence between English and five languages: Mandarin Chinese, Thai, and Vietnamese, which have lexical tone, and Japanese and Korean as additional comparisons. Correspondence is generally lower than previously reported European-language estimates. Adjusting for lexical tone increases pitch correspondence for English–Mandarin Chinese, English–Thai, and English–Vietnamese. However, English–Japanese and English–Korean also exhibit lower correspondence than the European-language estimates, indicating that tonal status alone does not explain the observed pattern. Greater word reordering is associated with weaker local correspondence even under semantic alignment, suggesting that semantic alignment does not fully account for differences in prosodic organization. These findings characterize the limits of fine-grained acoustic correspondence across translation and motivate prosody modeling that accounts for lexical pitch constraints and the structural context of translated content.

\end{abstract}

\input{Sections/1-introduction}

\input{Sections/2-related-work}

\input{Sections/3-method}
\input{Sections/4-results}
\input{Sections/5-conclusion}

\newpage

\section*{Limitations}

Licensing restrictions prevent us from sharing
the speech data, but our detailed pipeline
allows researchers to repeat the analyses
on comparable dubbed speech.
Acting, translation choices, and timing
constraints may limit generalization beyond
television dubbing.
Samples differ in size, content, speakers,
and production practices, especially for
direct tonal-language comparisons.
Using the same pipeline for French does
not remove these differences, so the observed
gaps cannot be explained by language
structure alone.

Automatic processing may introduce errors
that differ across languages.
Our gold alignment evaluations cover only
Chinese and Japanese text outside the dubbing
domain, so accuracy on the recordings remains
uncertain.
Coverage analyses and sensitivity checks
cannot rule out systematic errors or effects
of filtering.

Tone adjustment does not completely separate
lexical tone from intonation.
Detrending also removes meaningful pitch
patterns tied to sentence position.
The reordering analyses suggest relationships
rather than establish causes.
Higher correlations under temporal pairing
may reflect shared sentence-level patterns
without showing that corresponding words
carry similar prosody.

Listening studies are needed to determine
whether emphasis, emotion, and communicative
intent are preserved.
Episode-clustered confidence intervals may
not fully account for shared casts,
production practices, or English recordings
across language pairs.


\bibliography{custom,anthology}

\appendix

\input{Sections/appendix}

\end{document}

%% file: Sections/1-introduction.tex
\section{Introduction}
\label{sec:introduction}

Prosody conveys emphasis, emotion, and discourse structure beyond lexical content \citep{prosody_importance-Gussenhoven_2004, prosody_importance-article}. Preserving this information is an important goal of expressive speech-to-speech translation \citep{seamless-barrault2023seamless, prosody_transfer_study-swiatkowski2023cross, rasim-passt}. However, languages differ in how pitch encodes lexical distinctions and how translated content is ordered. Effective prosody transfer therefore requires understanding how these differences affect cross-lingual correspondence.

Professionally dubbed speech provides a useful setting for studying this correspondence. Dubbing adapts dialogue to another language while maintaining its role in the same scene and meeting expressive and temporal constraints \citep{brannon2023dubbing}. Using dubbed speech, \citet{xie2026prosodylosttranslationfinegrained} report positive word-aligned pitch and energy correlations for German--English, English--Spanish, and English--French. However, their study is limited to European-language pairs without lexical tone. It also reorders acoustic segments to align corresponding words, although lexical order is generally similar in the studied pairs. It therefore remains unclear whether these findings generalize to languages with lexical tone or substantial word reordering, and how much the measured correlations reflect local prosodic correspondence versus broader utterance-level patterns.


The first question concerns lexical tone. Pitch in tonal languages contributes to lexical distinctions as well as sentence-level prosody \citep{prosody_importance-Gussenhoven_2004, XU199955}. The lexical function of pitch might compromise the correspondence; semantically corresponding words may therefore have different pitch contours even when they serve similar expressive functions. If lexical tone contributes to lower measured correspondence, reducing tone-related variation should increase the similarity of aligned contours. Comparisons beyond tonal languages are also necessary to assess whether lower correspondence is specific to lexical tone. We therefore examine English paired with Mandarin Chinese, Thai, and Vietnamese, alongside Japanese and Korean. Japanese introduces a lexical pitch-accent system, while Korean provides an additional comparison beyond the three tonal languages.


The second question concerns how differences in sentence structure across languages relate to prosodic correspondence. Word-aligned analysis brings semantically corresponding units together, enabling local comparisons. However, language pairs with greater structural difference with different word-orders may exhibit greater reordering of corresponding regions, while language-specific segmentation may lead to lower alignment coverage. In such pairs, fewer regions may be aligned, and corresponding regions may occur in different positions within the utterance. Lower word-aligned correlation may therefore reflect both differences in prosodic realization and cross-language variation in word order and alignment coverage.


To our knowledge, this is the first systematic word-level study of prosodic correspondence across tonal and non-tonal Asian languages to examine both lexical tone and word reordering. We analyze pitch and energy for English paired with Mandarin Chinese, Thai, and Vietnamese, which have lexical tone, and with Japanese and Korean. We also compare all three tonal-language pairs directly. Episode-based cross-fitting estimates tone-related pitch variation, and matched-utterance comparisons measure the effect of adjustment. We assess alignment coverage and compare semantic and temporal pairing to examine how word reordering relates to measured correspondence, using a French reference processed with the same pipeline.

All examined pairs show positive mean pitch and energy correlations above unit-shuffled baselines. In contrast to the similar pitch correlations reported across three European-language pairs by \citet{xie2026prosodylosttranslationfinegrained}, we find lower pitch correspondence for all five English--Asian pairs. Energy correspondence above baseline is also lower. Tone adjustment increases pitch correspondence for English--Mandarin Chinese, English--Thai, and English--Vietnamese, with all gain intervals excluding zero, but adjusted means remain below both published European estimates and the current French reference. The gap therefore persists after tone adjustment and also occurs for Japanese and Korean, showing that the earlier findings do not generalize uniformly across languages and that tonal status alone does not explain the difference. Direct tonal-language comparisons show adjustment gains for Mandarin Chinese--Vietnamese and Vietnamese--Thai, but no clear change for Mandarin Chinese--Thai.


Japanese and Korean have the greatest mean reordering distances, lower alignment coverage, and the largest temporal-over-semantic pairing gains among reordered utterances. Greater reordering is associated with a larger temporal advantage for Vietnamese, Thai, Japanese, and Korean. Pitch correlation also tends to increase with alignment coverage, but coverage differences do not fully explain the lower correspondence: at 60–80\% coverage, French remains at 0.223, compared with 0.094–0.155 for the Asian-language pairs, and reweighting to the French coverage distribution does not close the gap. Semantic-pairing correlations remain positive across the Asian-language pairs, but are lower than the French reference. Together, these results highlight that word reordering and alignment coverage are important context for interpreting prosodic correspondence across language pairs.

Our contributions are: (1)~a systematic word-level analysis of prosodic correspondence across tonal and non-tonal Asian languages, including direct tonal-language comparisons and a European reference processed with the same pipeline; (2)~cross-fitted lexical-tone adjustment showing that tone-related variation contributes to pitch mismatch but tonal status alone does not explain the observed differences; and (3)~evidence that the degree of word reordering between different languages interacts with sentence-level pitch trends and semantic alignment to shape measured correspondence.

%% file: Sections/2-related-work.tex
\section{Related Work}
\label{sec:related}

\subsection{Prosody in Speech Translation}

Speech-to-speech translation has progressed from cascaded systems to direct models that jointly learn translation and speech generation
\citep{s2st_direct-jia2019direct,s2st_direct-jia2022translatotron,prosody_ignore-inaguma2023unity}.
Alongside improvements in translation quality and naturalness, several studies have explicitly addressed the preservation of expressive information.
Early work investigated the transfer of word-level emphasis
\citep{prosody_traditional_s2st_discard_prosody-7792638},
while subsequent approaches explored cross-lingual prosody transfer for expressive dubbing and multilingual speech translation
\citep{prosody_transfer_study-swiatkowski2023cross,seamless-barrault2023seamless}.
These approaches motivate a closer examination of which prosodic characteristics can be preserved across languages.

A complementary line of research examines whether translation systems use prosody to interpret source speech.
Studies of Korean wh-phrases and contrastive benchmarks show that direct speech-to-text translation models can exploit prosody, although their sensitivity remains limited
\citep{prosody_korean_study-zhou2024prosody,prosody_study-tsiamas2024speech}.
These studies concern the contribution of source prosody to translated meaning.
Our work instead examines the acoustic correspondence between source speech and human translations, providing empirical evidence relevant to the prosodic patterns that expressive S2ST systems aim to preserve.

\subsection{Cross-Lingual Prosodic Correspondence}

Human speech provides a complementary perspective on cross-lingual prosody.
\citet{brannon2023dubbing} analyze professional dubbing and find evidence that source audio influences translated speech beyond its lexical content.
Using bilingual conversational reenactments, \citet{avila2023dialog} investigate cross-language prosodic differences and develop a feature-based measure of prosodic dissimilarity.
These studies establish useful settings for examining how expressive characteristics are maintained or adapted across languages.

A recent study measures fine-grained pitch and energy correspondence in professionally dubbed English--German, English--Spanish, and English--French speech
\citep{xie2026prosodylosttranslationfinegrained}.
By comparing contours over aligned linguistic units, it finds cross-lingual correspondence above shuffled baselines.
However, its analysis is restricted to non-tonal European languages with similar structure.
We investigate prosodic correspondence involving languages with significant structural differences, such as Mandarin Chinese, Thai, and Vietnamese, where lexical tone adds a lexical function to pitch and sentence structures and word order may differ substantially.
We also examine sensitivity to broad utterance-level pitch trends and alignment choices, distinguishing these factors from the local correspondence measured by the analysis.

\subsection{Lexical Tone and Intonation}

Lexical tone and intonation jointly shape pitch in tonal languages \citep{prosody_importance-Gussenhoven_2004}: pitch distinguishes lexical meaning while also conveying prominence, phrasing, and communicative function. Studies document these interactions in Mandarin, where tone and focus jointly shape F0 contours \citep{XU199955}; Thai, where tonal realization varies by phrasal position \citep{82fcacd6-bb2f-3eb8-875f-e3f5a4466366}; and Vietnamese, where lexical tones interact with intonational patterns associated with communicative functions \citep{ha-viet-inproceedings}. Thus, tonal contours in connected speech reflect prosodic context as well as lexical patterns.

This motivates examining local pitch correspondence in translations involving tonal languages without assuming that lexical tone necessarily reduces correspondence. Whereas prior studies characterize tone–intonation interactions within individual languages, we examine how English source contours correspond to translated speech in three tonal targets. Comparing pitch and energy further tests whether cross-lingual differences are specific to pitch or extend to another acoustic dimension.

%% file: Sections/3-method.tex
\section{Data Processing and Analysis Pipeline}
\label{sec:methods}

We adapt the alignment-based analysis of
\citet{xie2026prosodylosttranslationfinegrained}
to Chinese, Vietnamese, Thai, Japanese,
and Korean.
The pipeline combines speech filtering,
timed transcription, bilingual alignment,
and prosodic contour comparison.
For Chinese, Vietnamese, and Thai, we
additionally estimate and remove tone-related
pitch variation.
Japanese and Korean provide further comparisons
for assessing whether lower correspondence
extends beyond these tonal languages.

\subsection{Data Extraction and Filtering}
\label{sec:data_filtering}

\paragraph{Speech Data.}
We analyze professionally dubbed television
dialogue mined using SpeechVecAlign
\citep{speechvecalign-meng2025speech}.
Each pair contains semantically corresponding
utterances extracted at their respective
track timestamps and converted to
16~kHz mono.
English serves as the common comparison
language, although the English recording
may itself be a dub.

We use all available Chinese, Vietnamese,
and Korean pairs, and sample 100,000 Thai
and 60,000 Japanese pairs.
We also process 10,000 English--French
pairs with the current settings.
The 2,713 retained pitch-scored pairs
provide a reference alongside the
published European-language results.

\paragraph{Audio and Semantic Filtering.}
Prosodic analysis can be sensitive to several factors, including audio quality, overlapping speech, multiple speakers within an utterance, and inconsistencies in semantic alignment. We therefore filter the dubbed data along these dimensions.  We apply SQUIM
\citep{squim-kumar2023torchaudio-squim}
to both recordings and retain pairs meeting
the thresholds in
Table~\ref{tab:filter_thresholds}.
We use pyannote diarization
\citep{pyannote-bredin2020pyannote}
to exclude pairs with more than one
detected speaker in either recording.
After transcription, we compute SONAR
cosine similarity between the transcripts
\citep{sonar-duquenne2023sonarsentencelevelmultimodallanguageagnostic}
and retain pairs scoring at least 0.45.
We repeat the analysis at 0.63 as a
semantic-filtering sensitivity test.
Tables~\ref{tab:filter_counts}
and~\ref{tab:corpus_statistics}
summarize the retained data.

\begin{table}[!htbp]
\centering
\small
\setlength{\tabcolsep}{4pt}
\begin{tabular*}{\columnwidth}{
@{\extracolsep{\fill}}lll@{}}
\toprule
\textbf{Stage}
& \textbf{Metric}
& \textbf{Threshold} \\
\midrule
Audio quality & SI-SDR & $\geq 7.5$ \\
Audio quality & STOI   & $\geq 0.91$ \\
Audio quality & PESQ   & $\geq 1.25$ \\
\midrule
Diarization & Speakers & $= 1$ \\
\midrule
Semantic filtering & SONAR cosine & $\geq 0.45$ \\
\bottomrule
\end{tabular*}
\caption{
Filtering thresholds.
Audio-quality and diarization criteria
apply to both recordings.}
\label{tab:filter_thresholds}
\vspace{-4mm}
\end{table}

\begin{table*}[t]
\centering
\small
\setlength{\tabcolsep}{4pt}
\begin{tabular*}{\textwidth}{
@{\extracolsep{\fill}}lrrrrrr@{}}
\toprule
\textbf{Stage}
& \textbf{EN--FR}
& \textbf{EN--ZH}
& \textbf{EN--VI}
& \textbf{EN--TH}
& \textbf{EN--JA}
& \textbf{EN--KO} \\
\midrule
Selected pairs
& 10,000 & 6,720 & 66,352 & 100,000 & 60,000 & 47,308 \\
Extracted pairs
& 9,994 & 6,720 & 66,352 & 99,932 & 60,000 & 47,308 \\
\midrule
After SQUIM
& 4,029 & 3,314 & 35,050 & 50,047 & 26,811 & 17,643 \\
After audio gates
& 2,923 & 2,641 & 26,332 & 38,342 & 19,775 & 14,127 \\
After SONAR
& 2,713 & 1,842 & 22,352 & 31,281 & 15,321 & 11,452 \\
\midrule
After pitch scoring
& 2,713 & 1,838 & 22,332 & 31,261 & 15,267 & 11,350 \\
\bottomrule
\end{tabular*}
\caption{
Utterance pairs retained at each processing
stage.
Audio gates include diarization.
Pitch scoring excludes pairs without valid,
nonconstant contour sequences.}
\label{tab:filter_counts}
\end{table*}

\begin{table}[!htbp]
\centering
\small
\setlength{\tabcolsep}{4pt}
\begin{tabular*}{\columnwidth}{
@{\extracolsep{\fill}}lrrr@{}}
\toprule
\textbf{Pair}
& \textbf{Utterances}
& \textbf{Episodes}
& \textbf{Hours} \\
\midrule
EN--FR & 2,713  & 1,324 & 5.4 \\
\midrule
EN--ZH & 1,838  & 55    & 2.6 \\
EN--VI & 22,332 & 446   & 37.8 \\
EN--TH & 31,261 & 1,132 & 54.5 \\
\midrule
EN--JA & 15,267 & 1,453 & 24.6 \\
EN--KO & 11,350 & 411   & 17.2 \\
\bottomrule
\end{tabular*}
\caption{
Final pitch-scored cohorts.
Approximate audio hours are summed across
both recordings.
French denotes the reference processed
with the current settings.}
\label{tab:corpus_statistics}
\vspace{-4mm}
\end{table}

\subsection{Transcription and Word Alignment}
\label{sec:transcription_alignment}

\paragraph{Timed Transcription.}
We transcribe each recording using
\texttt{whisper\_timestamped}
\citep{whisper_timestamped-lintoai2023whispertimestamped, whisper_timestamped-radford2022robust},
specifying the language of each recording.

\paragraph{Language-Specific Segmentation.}
We convert Chinese text to simplified
characters using OpenCC
\cite{opencc}
and segment with spaCy's
\texttt{zh\_core\_web\_sm}
\citep{spaCy-Honnibal_spaCy_Industrial-strength_Natural_2020}.
Japanese uses
\texttt{ja\_core\_news\_sm},
and Thai uses PyThaiNLP's
\texttt{newmm} tokenizer
\citep{phatthiyaphaibun-etal-2023-pythainlp}.
Vietnamese retains space-separated syllabic units.
For Korean, Kiwi morphemes
\cite{kiwi2} are used for alignment and mapped back
to the space-separated timed units
for acoustic extraction.

Timestamps are retained when segmentation
leaves a timed unit unchanged.
Only units requiring subdivision receive
estimated internal boundaries: their duration
is distributed across characters,
and each resulting segment receives
the span of its constituent characters.
The original timed unit determines the
outer boundaries of these segments.

\paragraph{Bilingual Word Alignment.}
We run FastAlign
\citep{fastalign-dyer2013simple}
in both directions with a diagonal prior
and retain intersecting links.
Training is augmented with TED parallel
text from OPUS
\citep{opus-TIEDEMANN12.463}:
TED2013 for English--Chinese and
English--French, and TED2020 for
the remaining English pairs.
Each background is capped at 143k
retained sentence pairs and tokenized
consistently with the speech transcripts.

\subsection{Prosodic Contour Construction}
\label{sec:prosody_representation}

\paragraph{Pitch and Energy.}
We extract pitch using YAAPT through
pYAAPT
\citep{pyaapt-kasi2002yet}.
Unvoiced regions are interpolated to
obtain continuous contours.
Energy is computed as RMS amplitude
computed with librosa
\citep{librosa-mcfee2015librosa},
using a 25~ms window and 10~ms hop.
Values below the fifth percentile
are set to zero.

\paragraph{Alignment Clusters.}
We group transcript units connected by bilingual
alignment links into clusters, rather than grouping
words by adjacency.
A cluster may contain one or several units
on either side.
Using their timestamps, we extract each side's
acoustic frames and resample them to 20 frames.
We then concatenate paired chunks in the order
of each cluster's first English-side unit,
placing corresponding content at matching
positions despite differences in word order.

We normalize positive feature values by their utterance-level mean to reduce speaker-dependent effects while preserving prosodic structure, since prosodic prominence is often conveyed relative to surrounding words and phrases.
Pairs with constant or otherwise invalid
score sequences are excluded.
Contour-resolution and track-quality
checks are reported in
Appendix~\ref{app:feature_checks}.

\subsection{Correlation Analysis}
\label{sec:correlation_method}

\paragraph{Correlation Metric and Baseline.}
For each utterance, we compute zero-lag
Spearman rank correlation \citep{spearman}
between the paired contour sequences
and report the mean across finite scores. Prosodic patterns are meaningful in context, so we focus measuring the correspondence of the relative ordering of values rather than their absolute magnitudes.
Pitch and energy means use their respective
valid cohorts.
Paired comparisons retain utterances
with valid scores in both conditions.

The shuffled baseline randomly
permutes target clusters once
within each utterance before scoring.
We report its mean alongside the observed
correlation and calculate correspondence
above baseline from paired utterance-level
differences.

\paragraph{Statistical Analysis.}
We estimate 95\% confidence intervals
using 2,000 episode-clustered bootstrap
draws.
For paired contrasts, resampling retains
the pairing between conditions.
Reported changes are calculated from
utterance-level differences rather than
subtraction of
corpus means.

\subsection{Lexical Tone Adjustment}
\label{sec:tone_method}

We assess the contribution of lexical
tone by estimating tone-related pitch
variation and repeating the contour
comparison after normalizing and adjusting according to the estimated lexical tone.

\paragraph{Tone Labels and Timing.}
We derive syllable-level tone labels
using pypinyin for Mandarin
\citep{pypinyin},
orthographic rules and pronunciation
exceptions for Thai, and written tone
markings for Vietnamese.
MMS forced alignment
\citep{mms1,mms2}
locates romanized syllables in the audio.
Successive syllable onsets define the
intervals used by the tone model.

\paragraph{Tone Model.}
We estimate typical pitch patterns from each
syllable's tone label and those of adjacent
syllables not separated by a pause, using
ridge-regularized linear regression.
Pitch is expressed in semitones relative
to the utterance's median F0.
The model represents each tone's contribution
as a smooth curve over syllable time
that varies with syllable duration.
This captures both a high tone's pitch offset
and a falling tone's changing contour.
The model also includes tone-pair interactions,
utterance position, and pauses.
We subtract only the centered tone and
tone-context contributions from the analysis
pitch track, retaining the shared syllable
contour and modeled utterance-level trends.

\paragraph{Cross-Fitted Adjustment.}
We divide episodes into five groups,
fit the model on four groups, and
adjust the remaining group.
Each utterance is therefore adjusted
using a model that has not seen
its episode.
We subtract the estimated component
in semitone space, convert the adjusted
track back to Hz, and repeat the
same contour construction and scoring
procedure used for raw pitch.
The primary adjustment effect is
the paired adjusted-minus-raw change
in correlation on matched utterances.

Validation compares adjustment with
real and shuffled tone labels and
measures tone predictability before
and after adjustment
(Appendix~\ref{app:tone_validation}).
We refer to the resulting tracks
as tone-adjusted pitch.

\subsection{Direct Tonal-Language Comparisons}
\label{sec:tonal_pair_method}

We identify Chinese--Vietnamese,
Chinese--Thai, and Vietnamese--Thai
recordings through matching English
recordings and English
segment boundaries.
Each recording retains its own timestamps.
We apply SONAR filtering directly
between the two dub transcripts
and train direct bilingual alignments
with paired TED2020 backgrounds.

We retain utterances with finite pitch
scores for the direct tonal comparison
and both corresponding English comparisons.
Results are reported for the direct
tonal pair, with clusters ordered
by the first language.
We adjust both recordings and compare
raw and adjusted correlations on matched
utterances, yielding 425
Chinese--Vietnamese, 279 Chinese--Thai,
and 3,441 Vietnamese--Thai pairs.

\subsection{Alignment Coverage and Cluster Ordering}
\label{sec:alignment_method}

We examine how coverage and aligned cluster
ordering affect measured correspondence,
using the French reference processed
with the current settings. We define a cluster as a semantically aligned unit that may comprise one or more words in either language.

\paragraph{Coverage and Reordering.}
We measure coverage as the proportion
of English words linked by FastAlign
and the proportion of English speech
time within analyzed clusters.
English speech-time coverage provides
a common measure across different
target-language segmentations.
We compare correlations within coverage
bins and reweight their contributions
to match the French coverage.

For utterances with at least two
clusters, we measure reordering using
normalized Kendall distance
\cite{kendall}
between the English and target cluster orders.
A distance of zero indicates identical
order and one complete reversal.

\paragraph{Semantic and Temporal Pairing.}
The main analysis uses semantic pairing:
target clusters follow the order of
their English counterparts.
We compare this with temporal pairing,
in which the same target clusters
retain their own chronological order.
The English sequence, selected clusters,
acoustic frames, and resampling remain
fixed across conditions.

We report mean correlations under
both pairing conditions for utterances
with at least two aligned clusters.
We also calculate the paired
temporal-minus-semantic correlation
difference among reordered utterances,
defined as those with nonzero
reordering distance.
When cluster orders agree, the two
pairing conditions are identical.
Detailed coverage and ordering analyses
appear in Appendix~\ref{app:alignment_sensitivity}.


\subsection{Temporal Feature Analysis}
\label{sec:temporal_method}

We measure utterance duration from
timed transcript extent and derive phoneme counts using
Phonemizer \citep{bernard2021phonemizer}
with eSpeak NG
\citep{espeakng},
and TLTK for Thai
\citep{tltk}.
We retain pairs with valid outputs
and equal positive sentence counts.

Duration and phoneme-count Spearman
correlations are computed across utterances,
separately from the within-utterance
contour correlations.
The phoneme counts represent canonical
pronunciations derived from recognized
text.

\subsection{Contributions by Part of Speech}
\label{sec:pos_method}
We assign alignment clusters to nouns, verbs, modifiers, or function words using their dominant English-side spaCy POS tags \citep{spaCy-Honnibal_spaCy_Industrial-strength_Natural_2020}. For each group, its contribution is the full correlation minus the correlation after removing that group's frames from both contours; positive values indicate reduced correspondence. We repeat this analysis for raw and tone-adjusted pitch. Further methodological details, including the French reference sample and the non-additivity of contributions, are provided in Appendix~\ref{app:pos_results}.



%% file: Sections/4-results.tex
\section{Prosody Analysis Results}
\label{sec:results}

\begin{figure}[t]
\centering
\includegraphics[width=\columnwidth]
    {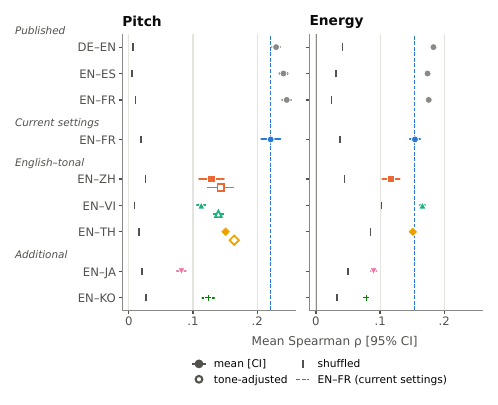}
\caption{
Mean pitch and energy correlations with
95\% confidence intervals and shuffled
baselines.
Filled markers indicate raw correlations;
open markers indicate tone-adjusted pitch.
Vertical ticks mark baseline means.}
\label{fig:main_correlations}
\vspace{-4mm}
\end{figure}

\begin{table*}[t]
\centering
\scriptsize
\setlength{\tabcolsep}{2pt}
\begin{tabular*}{\textwidth}{
@{\extracolsep{\fill}}lrcccccc@{}}
\toprule
& & \multicolumn{2}{c}{\textbf{Raw pitch}}
& \multicolumn{2}{c}{\textbf{Pitch adjustment}}
& \multicolumn{2}{c}{\textbf{Energy}} \\
\cmidrule(lr){3-4}
\cmidrule(lr){5-6}
\cmidrule(lr){7-8}
\textbf{Pair}
& \textbf{$n$}
& \textbf{Mean $\rho$ [95\% CI]}
& \textbf{Baseline}
& \textbf{Adj.\ $\rho$ [95\% CI]}
& \textbf{$\Delta_{\mathrm{adjust}}$ [95\% CI]}
& \textbf{Mean $\rho$ [95\% CI]}
& \textbf{Baseline} \\
\midrule
\multicolumn{8}{l}{
\textit{Published results:
\citet{xie2026prosodylosttranslationfinegrained}}} \\
DE--EN & 16,335
& 0.229 [0.223, 0.236] & 0.007
& --- & ---
& 0.183 [0.179, 0.187] & 0.043 \\
EN--ES & 16,939
& 0.241 [0.235, 0.247] & 0.005
& --- & ---
& 0.174 [0.170, 0.178] & 0.030 \\
EN--FR & 15,957
& 0.246 [0.239, 0.253] & 0.010
& --- & ---
& 0.176 [0.172, 0.179] & 0.026 \\
\midrule
\multicolumn{8}{l}{
\textit{Current study: same-pipeline reference}} \\
EN--FR & 2,713
& 0.221 [0.206, 0.236] & 0.019
& --- & ---
& 0.154 [0.146, 0.162] & 0.038 \\
\midrule
\multicolumn{8}{l}{
\textit{Current study: English--tonal pairs}} \\
EN--ZH & 1,838
& 0.129 [0.110, 0.148] & 0.026
& 0.143 [0.123, 0.163]
& +0.015 [0.003, 0.025]
& 0.116 [0.103, 0.130] & 0.045 \\
EN--VI & 22,332
& 0.113 [0.106, 0.119] & 0.009
& 0.140 [0.132, 0.147]
& +0.027 [0.023, 0.030]
& 0.166 [0.162, 0.170] & 0.102 \\
EN--TH & 31,261
& 0.151 [0.146, 0.155] & 0.016
& 0.164 [0.160, 0.169]
& +0.013 [0.012, 0.015]
& 0.151 [0.148, 0.153] & 0.085 \\
\midrule
\multicolumn{8}{l}{
\textit{Current study: Japanese and Korean comparisons}} \\
EN--JA & 15,267
& 0.082 [0.075, 0.089] & 0.020
& --- & ---
& 0.090 [0.086, 0.094] & 0.050 \\
EN--KO & 11,350
& 0.124 [0.115, 0.133] & 0.026
& --- & ---
& 0.078 [0.073, 0.083] & 0.033 \\
\midrule
\multicolumn{8}{l}{
\textit{Current study: tonal--tonal pairs}} \\
ZH--VI & 425
& 0.063 [0.023, 0.103] & 0.001
& 0.096 [0.058, 0.130]
& +0.033 [0.006, 0.056]
& 0.130 [0.113, 0.151] & 0.075 \\
ZH--TH & 279
& 0.149 [0.106, 0.190] & 0.006
& 0.149 [0.104, 0.193]
& +0.000 [$-0.026$, 0.023]
& 0.092 [0.072, 0.111] & 0.052 \\
VI--TH & 3,441
& 0.088 [0.076, 0.100] & 0.007
& 0.150 [0.137, 0.163]
& +0.062 [0.052, 0.072]
& 0.190 [0.182, 0.197] & 0.136 \\
\bottomrule
\end{tabular*}
\caption{
Mean utterance-level correlations with
95\% confidence intervals and unit-shuffled
baselines.
Current-study intervals use episode-clustered
bootstrapping.
The current French reference uses the
same processing settings as the new
language pairs.
$\Delta_{\mathrm{adjust}}$ is the paired
adjusted-minus-raw pitch change.
For English--tonal pairs, $n$ denotes
the raw cohort.}
\label{tab:main_correlations}
\vspace{-4mm}
\end{table*}

\subsection{English--Tonal-Language Comparisons}
\label{sec:main_correlations}



We first compare English with the three languages that use lexical tone. Raw pitch correlations are lower for English--Mandarin Chinese (0.129), English--Vietnamese (0.113), and English--Thai (0.151) than for the English--French reference (0.221). The differences are substantial: even the highest English--tonal correlation is below the French value.
We then estimate how much of this gap is associated with lexical tone by adjusting pitch for tone-related variation. The correlations increase to 0.143 for Chinese, 0.140 for Vietnamese, and 0.164 for Thai. These gains are positive but modest (+0.015, +0.027, and +0.013, respectively), and all three adjusted correlations remain below the French reference. Tone-related pitch variation therefore contributes to the difference, but adjusting for it does not close the gap. Lexical tone alone cannot account for the lower local pitch correspondence between English and these languages.
Energy correlations show a different pattern: they are 0.116 for Chinese, 0.166 for Vietnamese, and 0.151 for Thai, compared with 0.154 for French. Thus, the lower correlations are not uniform across acoustic measures. The pitch results motivate examining lexical tone, while the energy results caution against treating the difference as a general reduction in all prosodic correspondence.



\subsection{Tonal--Tonal Comparisons}

The three tonal-language pairs show that sharing a lexical tone structure does not by itself yield strong local pitch correspondence. Raw correlations range from 0.063 for Chinese--Vietnamese to 0.149 for Chinese--Thai, with Vietnamese--Thai at 0.088. Tone adjustment increases the Chinese--Vietnamese correlation to 0.096 and the Vietnamese--Thai correlation to 0.150, but leaves the Chinese--Thai correlation essentially unchanged at 0.149. The effect of adjustment therefore varies across pairs, and the resulting correlations remain modest.

These results suggest that lexical tone contributes to cross-lingual pitch differences, but does not account for them on its own. Even when both languages use lexical tone, their local pitch patterns do not necessarily correspond. Other differences between the languages and between the structures of translated utterances must also be considered.

\subsection{English--Japanese and English--Korean Comparisons}

The English--Japanese and English--Korean pairs provide a further comparison with languages that do not use lexical tone. Japanese uses lexical pitch accent, while Korean does not use lexical tone. Their raw pitch correlations with English are 0.082 and 0.124, respectively; their energy correlations are 0.090 and 0.078. The low pitch correlations for these pairs show that lexical tone is not necessary for weak cross-lingual local pitch correspondence. Taken together with the tonal-language results, this comparison indicates that tone status alone does not explain the observed differences.

These pairs also differ more strongly from English in the ordering of semantically aligned words and have lower alignment coverage. This co-occurrence motivates examining syntactic word-order divergence as an indicator of broader differences in sentence structure. 

\begin{table*}[t]
\centering
\small
\setlength{\tabcolsep}{4pt}
\begin{tabular*}{\textwidth}{
@{\extracolsep{\fill}}lrccc@{}}
\toprule
\textbf{Pair}
& \textbf{Mean $R$}
& \textbf{Semantic $\rho$}
& \textbf{Temporal $\rho$}
& \textbf{Temporal advantage [95\% CI]} \\
\midrule
EN--FR & 0.045 & 0.221 & 0.227
& +0.013 [$-0.001$, 0.025] \\
EN--ZH & 0.094 & 0.132 & 0.144
& +0.025 [0.001, 0.049] \\
EN--VI & 0.063 & 0.113 & 0.121
& +0.015 [0.010, 0.021] \\
EN--TH & 0.072 & 0.151 & 0.169
& +0.033 [0.029, 0.037] \\
EN--JA & 0.241 & 0.083 & 0.141
& +0.090 [0.080, 0.101] \\
EN--KO & 0.151 & 0.130 & 0.183
& +0.116 [0.103, 0.129] \\
\bottomrule
\end{tabular*}
\caption{
Cluster reordering and raw-pitch correspondence
under semantic and temporal pairing.
$R$ is normalized Kendall distance between
cluster orders, ranging from zero for identical
order to one for complete reversal.
Mean $R$ and mean correlations use utterances
with at least two aligned clusters; this
restriction accounts for differences from
Table~\ref{tab:main_correlations}.
Temporal advantage is the mean paired
$\rho_{\mathrm{temporal}}-\rho_{\mathrm{semantic}}$
among reordered utterances ($R>0$), rather than
the difference between the displayed corpus means.
Intervals use episode-clustered bootstrapping.
}
\label{tab:pairing_sensitivity}
\end{table*}

\subsection{Word Reordering and Alignment Coverage}

The mean normalized reordering distance is greatest for Japanese (0.241) and Korean (0.151), compared with French (0.045) and the tonal-language pairs (0.063--0.094). Alignment coverage follows a similar broad pattern: English speech-time coverage is 52\% for Japanese and 46\% for Korean, compared with 62--66\% for the tonal-language pairs and 72\% for French. Word coverage is also lower for Japanese and Korean (42\% and 39\%) than for French (67\%) or the tonal-language pairs (55--60\%). These measures show that translated utterances differ not only in the ordering of aligned content but also in how much content can be aligned.

We compare semantic pairing, which brings aligned words together, with chronological pairing, which preserves their original order in each recording. The displayed semantic-pairing pitch correlations are 0.221 for French, 0.132 for Chinese, 0.113 for Vietnamese, 0.151 for Thai, 0.083 for Japanese, and 0.130 for Korean. For Japanese and Korean, the corresponding chronological-pairing correlations are higher: 0.141 and 0.183. Among reordered utterances, the mean paired changes from semantic to chronological pairing are +0.090 for Japanese and +0.116 for Korean. The changes are smaller for French (+0.013), Chinese (+0.025), Vietnamese (+0.015), and Thai (+0.033).

For Japanese and Korean, then, semantic reordering substantially lowers the measured correlation relative to preserving the original sentence order. Aligning words by meaning does not recover strong local prosodic correspondence in these pairs. Instead, it produces much weaker correlations, consistent with the possibility that semantically corresponding words occur in different prosodic contexts when sentence structures differ. The higher correlations under chronological pairing suggest that some broader sentence-level pitch organization remains shared; they do not show that local prosody corresponds at the semantic word level.


\subsection{Summary}
Tone adjustment yields modest pitch-correspondence gains for English--tonal pairs and larger gains for two tonal--tonal pairs, but correlations remain modest. Low correspondence also occurs for English--Japanese and English--Korean, indicating that the pattern is not specific to lexical tone. Japanese and Korean show greater word-order divergence and lower alignment coverage, and their correlations are substantially higher under temporal than semantic pairing. Their greater word-order divergence indicates greater syntactic differences, while the low semantic-pairing correlations suggest that fine-grained prosodic correspondence may be weaker across structurally different language pairs. The higher temporal-pairing correlations may reflect shared utterance-level pitch structure rather than correspondence between semantically aligned words. Appendices~\ref{app:pos_results} and~\ref{app:temporal_results} provide complementary analyses of POS contributions and utterance duration correspondence.

%% file: Sections/5-conclusion.tex
\section{Conclusion}
\label{sec:conclusion}

We analyzed word-level prosodic correspondence in professionally dubbed speech across five English--target language pairs and three pairs among tonal languages. All pairs show positive mean pitch and energy correlations above unit-shuffled baselines. Pitch correspondence for the English--target pairs remains below European-language estimates, while energy shows a more mixed pattern. Tone adjustment improves all three English--tonal pitch correlations but does not close the gap; low correlations for English--Japanese and English--Korean also show that weak correspondence is not specific to lexical tone.

Japanese and Korean show the greatest word-order divergence and lowest alignment coverage. Their correlations are substantially higher under temporal than semantic pairing, indicating that semantic alignment does not recover local prosodic correspondence in these pairs. The greater word-order divergence reflects broader syntactic differences, which are associated with weaker fine-grained prosodic correspondence. The higher correlations under temporal pairing may instead reflect shared utterance-level pitch structure.

Together, our findings indicate that lexical tone contributes to cross-lingual pitch differences, but does not account for them on its own. Local prosodic correspondence may be weak across languages with substantially different syntactic structures, even when some broader utterance-level correspondence remains. This motivates expressive speech translation approaches that consider both sentence structure and their prosodic context.

%% file: Sections/appendix.tex
\section{Lexical Tone Adjustment Validation}
\label{app:tone_validation}

\subsection{Tone-Label-Shuffled Baseline}
\label{app:tone_label_baseline}

The main results report paired changes
from raw to tone-adjusted pitch.
Here, we test whether the gains depend
on the tone labels by fitting the
same adjustment model with shuffled
labels, using the same episode-based
cross-fitting procedure.
We denote the paired correlation
difference between real-label and
shuffled-label adjustment by
$\Delta_{\mathrm{label}}$.
This baseline evaluates the adjustment
procedure; the unit-shuffled baseline
instead permutes target alignment
clusters before correlation is measured.

Real-label adjustment exceeds the
tone-label-shuffled baseline by 0.016
for Chinese, 0.027 for Vietnamese,
and 0.014 for Thai, with all three
confidence intervals excluding zero
(Table~\ref{tab:tone_label_baseline}).
The gains therefore depend on information
carried by the tone labels.
This comparison supports the specificity
of the adjustment, without requiring
a complete separation of tone and
intonation.

\begin{table}[H]
\centering
\small
\setlength{\tabcolsep}{4pt}
\begin{tabular*}{\columnwidth}{
@{\extracolsep{\fill}}lc@{}}
\toprule
\textbf{Pair}
& \textbf{$\Delta_{\mathrm{label}}$ [95\% CI]} \\
\midrule
EN--ZH & +0.016 [0.005, 0.026] \\
EN--VI & +0.027 [0.024, 0.030] \\
EN--TH & +0.014 [0.012, 0.016] \\
\bottomrule
\end{tabular*}
\caption{
Paired differences between real-label
and shuffled-label tone adjustment on
matched eligible utterances.
Intervals use episode-clustered bootstrapping.
These contrasts differ from the
adjusted-minus-raw changes in the
main results.}
\label{tab:tone_label_baseline}
\end{table}

\subsection{Tonal-Pair Label Baseline}
\label{app:tonal_pair_baseline}

For the direct tonal-language pairs,
we compare adjustment of both recordings
using real labels with adjustment of
both recordings using shuffled labels.
The point differences are 0.035 for
Chinese--Vietnamese, 0.005 for
Chinese--Thai, and 0.063 for
Vietnamese--Thai
(Table~\ref{tab:tonal_pair_label_baseline}).
Their ordering agrees with the
adjusted-minus-raw changes reported
in Table~\ref{tab:main_correlations}.

\begin{table}[H]
\centering
\small
\setlength{\tabcolsep}{4pt}
\begin{tabular*}{\columnwidth}{
@{\extracolsep{\fill}}lrc@{}}
\toprule
\textbf{Pair}
& \textbf{Matched $n$}
& \textbf{$\Delta_{\mathrm{label}}$} \\
\midrule
ZH--VI & 425   & +0.035 \\
ZH--TH & 279   & +0.005 \\
VI--TH & 3,441 & +0.063 \\
\bottomrule
\end{tabular*}
\caption{
Real-label-minus-shuffled-label pitch
correlation differences, with both
recordings adjusted on matched utterances.
Only point estimates are reported for
these contrasts; the adjusted-minus-raw
intervals in the main results refer
to a different comparison.}
\label{tab:tonal_pair_label_baseline}
\end{table}

\subsection{Tone Contours and Predictability}
\label{app:tone_predictability}

\begin{figure*}[t]
\centering
\includegraphics[width=\textwidth]
    {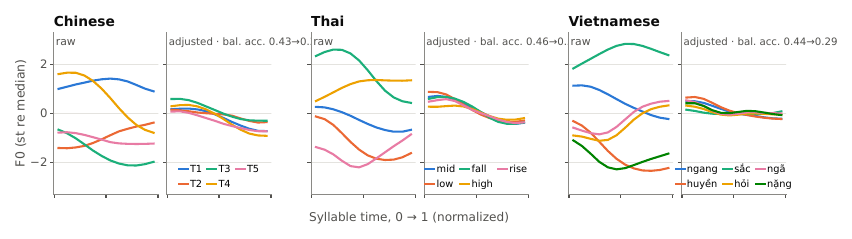}
\caption{
Tone-conditioned syllable pitch contours
before and after adjustment for Chinese,
Thai, and Vietnamese.
Pitch is expressed in semitones relative
to the utterance median, with syllable
time normalized from zero to one.
Colors identify tone categories within
each language.
The annotations report tone-classification
balanced accuracy before and after
adjustment.}
\label{fig:tone_contours}
\end{figure*}

Figure~\ref{fig:tone_contours} compares
tone-conditioned syllable contours
before and after adjustment.
Raw contours show differences in
pitch level and shape across tone
categories.
These differences are reduced after
adjustment, although some separation
between categories remains.

We also measure tone predictability
from syllable contours using the
3,000-utterance diagnostic sample.
Balanced accuracy decreases from
0.427 to 0.301 for Chinese, from
0.444 to 0.287 for Vietnamese, and
from 0.464 to 0.333 for Thai
(Table~\ref{tab:tone_classification}).
These changes correspond to reductions
of approximately 55\%, 57\%, and
50\% in above-chance balanced accuracy.
Adjusted accuracy remains above chance
in all three languages.
The adjustment therefore reduces tone
predictability without eliminating it;
the percentages describe reductions
in classification performance, rather
than fractions of tone information
removed.

\begin{table}[H]
\centering
\small
\setlength{\tabcolsep}{4pt}
\begin{tabular*}{\columnwidth}{
@{\extracolsep{\fill}}lccc@{}}
\toprule
\textbf{Language}
& \textbf{Raw}
& \textbf{Adjusted}
& \textbf{Chance} \\
\midrule
Chinese    & 0.427 & 0.301 & 0.200 \\
Vietnamese & 0.444 & 0.287 & 0.167 \\
Thai       & 0.464 & 0.333 & 0.200 \\
\bottomrule
\end{tabular*}
\caption{
Tone-classification balanced accuracy
from raw and adjusted syllable contours
on the diagnostic sample.}
\label{tab:tone_classification}
\end{table}

\section{Alignment and Reordering Sensitivity}
\label{app:alignment_sensitivity}

\subsection{Alignment Coverage}
\label{app:alignment_coverage}

Table~\ref{tab:alignment_coverage}
reports coverage and cluster ordering
for the raw-pitch runs.
French uses the current processing
settings throughout this section.
FastAlign links 67\% of English words
for French, compared with 55--60\%
for the three tonal languages,
42\% for Japanese, and 39\% for Korean.
English speech-time coverage follows
the same pattern.
Japanese and Korean also have fewer
clusters per utterance, averaging
4.1 and 3.4, respectively, compared
with 8.0 for French.

\begin{table*}[t]
\centering
\small
\setlength{\tabcolsep}{4pt}
\begin{tabular*}{\textwidth}{
@{\extracolsep{\fill}}lrrrrr@{}}
\toprule
& \multicolumn{2}{c}{\textbf{English coverage}}
& & \multicolumn{2}{c}{\textbf{Reordering}} \\
\cmidrule(lr){2-3}
\cmidrule(lr){5-6}
\textbf{Pair}
& \textbf{Words}
& \textbf{Speech time}
& \textbf{Clusters/utt.}
& \textbf{Any}
& \textbf{Mean $R$} \\
\midrule
EN--FR & 67\% & 72\% & 8.0 & 48\% & 0.045 \\
EN--ZH & 55\% & 62\% & 5.2 & 48\% & 0.094 \\
EN--VI & 60\% & 66\% & 6.4 & 50\% & 0.063 \\
EN--TH & 59\% & 66\% & 6.4 & 53\% & 0.072 \\
EN--JA & 42\% & 52\% & 4.1 & 64\% & 0.241 \\
EN--KO & 39\% & 46\% & 3.4 & 46\% & 0.151 \\
\bottomrule
\end{tabular*}
\caption{
Alignment coverage and cluster reordering
in the raw-pitch runs.
Word coverage is the proportion of
English words linked by FastAlign;
speech-time coverage is the proportion
of English speech time within analyzed
clusters.
Reordering statistics use utterances
with at least two clusters.
``Any'' denotes the proportion with
$R>0$, where $R$ is normalized Kendall
distance between cluster orders.
Zero indicates identical order and
one complete reversal.}
\label{tab:alignment_coverage}
\end{table*}

Pitch correlation generally increases
with coverage.
For Japanese, the mean rises from
0.070 in the lowest coverage bin
to 0.115 in the highest.
At 60--80\% English speech-time
coverage, French reaches 0.223,
compared with 0.094--0.155 for
the Asian-language pairs.
Reweighting each pair to the French
coverage distribution increases the
Japanese and Korean means by approximately
0.015 and the tonal means by less
than 0.005.
These point estimates show that matching
the coverage distribution does not
eliminate the difference.

\subsection{Cluster Reordering}
\label{app:cluster_reordering}

We measure reordering by comparing
the English order of aligned clusters
with their order in the target recording.
The frequency of any reordering and
its extent capture different properties:
French and Korean have similar proportions
of reordered utterances, at 48\% and
46\%, but mean distances of 0.045
and 0.151, respectively.
Japanese has the largest mean distance,
at 0.241, compared with 0.063--0.094
for the English--tonal pairs.

These statistics describe automatically
recovered alignments.
Missed links can suppress observed
crossings, while incorrect links can
introduce them.
Utterances with fewer than two clusters
are excluded from the reordering analysis;
12\% of Korean utterances contain
only one cluster.

\subsection{Semantic and Temporal Pairing}
\label{app:pairing_sensitivity}

We compare two pairings of the same
aligned clusters.
Semantic pairing places target clusters
in the order of their English counterparts,
as in the main analysis.
Temporal pairing retains the target
clusters' chronological order while
leaving the English sequence fixed.
The selected clusters, acoustic frames,
and resampling procedure are identical,
so this comparison changes ordering
while holding coverage constant.

We define the temporal advantage as
$D=\rho_{\mathrm{temporal}}
-\rho_{\mathrm{semantic}}$.
Table~\ref{tab:pairing_sensitivity} in the
main results reports raw correlations and
paired differences.
Table~\ref{tab:pairing_detrending} compares
the paired differences before and after
detrending among reordered utterances
($R>0$) with at least two clusters.
The pairing conditions coincide when
the cluster orders agree.

\begin{table}[t]
\centering
\small
\setlength{\tabcolsep}{4pt}
\begin{tabular*}{\columnwidth}{
@{\extracolsep{\fill}}lcc@{}}
\toprule
& \multicolumn{2}{c}{
\textbf{Temporal advantage $D$ [95\% CI]}} \\
\cmidrule(lr){2-3}
\textbf{Pair}
& \textbf{Before detrending}
& \textbf{After detrending} \\
\midrule
EN--FR
& +0.013 [$-0.001$, 0.025]
& +0.000 [$-0.015$, 0.015] \\
EN--ZH
& +0.025 [0.001, 0.049]
& +0.007 [$-0.017$, 0.032] \\
EN--VI
& +0.015 [0.010, 0.021]
& +0.007 [0.001, 0.013] \\
EN--TH
& +0.033 [0.029, 0.037]
& +0.018 [0.013, 0.023] \\
EN--JA
& +0.090 [0.080, 0.101]
& +0.005 [$-0.004$, 0.014] \\
EN--KO
& +0.116 [0.103, 0.129]
& +0.023 [0.010, 0.035] \\
\bottomrule
\end{tabular*}
\caption{
Paired temporal-minus-semantic pitch correlation
differences among reordered utterances ($R>0$)
with at least two aligned clusters, before and
after removing each recording's linear pitch trend.
Both conditions use the same selected clusters.
Intervals use episode-clustered bootstrapping.
French uses the current processing settings.}
\label{tab:pairing_detrending}
\end{table}

Japanese and Korean have the largest
paired temporal advantages, at 0.090
and 0.116, respectively.
The gains are 0.025 for Chinese,
0.015 for Vietnamese, 0.033 for Thai,
and 0.013 for French.
All intervals except the French
interval exclude zero.

\subsection{Within-Utterance Pitch Trends}
\label{app:pitch_trends}

Table~\ref{tab:pitch_trends} provides the
full pitch-trend statistics summarized
in Table~\ref{tab:detrended_correlations}.
Each recording's trend is fitted in
its own temporal cluster order, using
utterances with at least three clusters.
The per-cluster slope depends on the
number of clusters, whereas the fitted
first-to-last change summarizes the
trend across the analyzed utterance.

Mean first-to-last changes are negative
on both sides of every language pair,
with all corresponding confidence
intervals excluding zero.
Between 65\% and 76\% of utterances
have falling raw-pitch trends.
Declining trends also remain after
tone adjustment, with mean first-to-last
changes of $-10.7\%$ for Chinese,
$-6.8\%$ for Vietnamese, and $-14.0\%$
for Thai.
Tone adjustment therefore retains
broad sentence-level pitch decline.

\begin{table*}[!t]
\centering
\scriptsize
\setlength{\tabcolsep}{2pt}
\begin{tabular*}{\textwidth}{
@{\extracolsep{\fill}}lrrlcccrr@{}}
\toprule
& & & &
& \multicolumn{2}{c}{
\textbf{Fitted first-to-last change}}
& & \\
\cmidrule(lr){6-7}
\textbf{Pair}
& \textbf{$n$}
& \shortstack{\textbf{Mean}\\\textbf{clusters}}
& \textbf{Recording}
& \shortstack{\textbf{Slope per cluster}\\
(\% of mean $F_0$)}
& \shortstack{\% of mean $F_0$}
& \textbf{Semitones}
& \shortstack{\textbf{Falling}\\(\%)}
& \textbf{$R^2$} \\
\midrule
EN--FR & 2,595 & 8.3 & English
& $-2.9$ [$-3.1$, $-2.6$]
& $-14.5$ [$-15.7$, $-13.4$]
& $-2.58$ [$-2.78$, $-2.39$]
& 76 & 0.42 \\
& & & French
& $-2.5$ [$-2.8$, $-2.2$]
& $-11.8$ [$-13.0$, $-10.6$]
& $-2.13$ [$-2.33$, $-1.92$]
& 72 & 0.41 \\
\midrule
EN--ZH & 1,627 & 5.6 & English
& $-3.5$ [$-4.0$, $-3.0$]
& $-12.3$ [$-13.7$, $-10.7$]
& $-2.16$ [$-2.43$, $-1.89$]
& 70 & 0.48 \\
& & & Chinese
& $-3.0$ [$-3.4$, $-2.6$]
& $-10.1$ [$-11.3$, $-8.9$]
& $-1.81$ [$-2.03$, $-1.59$]
& 68 & 0.41 \\
& & & Chinese (adjusted)
& $-3.3$ [$-3.6$, $-2.9$]
& $-10.7$ [$-11.9$, $-9.5$]
& $-1.92$ [$-2.13$, $-1.71$]
& 70 & 0.45 \\
\midrule
EN--VI & 20,590 & 6.7 & English
& $-3.3$ [$-3.4$, $-3.2$]
& $-14.1$ [$-14.5$, $-13.6$]
& $-2.50$ [$-2.57$, $-2.42$]
& 76 & 0.46 \\
& & & Vietnamese
& $-1.7$ [$-1.8$, $-1.5$]
& $-6.8$ [$-7.3$, $-6.4$]
& $-1.23$ [$-1.31$, $-1.15$]
& 65 & 0.32 \\
& & & Vietnamese (adjusted)
& $-1.6$ [$-1.8$, $-1.5$]
& $-6.8$ [$-7.2$, $-6.4$]
& $-1.21$ [$-1.29$, $-1.13$]
& 69 & 0.38 \\
\midrule
EN--TH & 28,817 & 6.8 & English
& $-3.3$ [$-3.4$, $-3.2$]
& $-13.8$ [$-14.2$, $-13.5$]
& $-2.44$ [$-2.51$, $-2.39$]
& 76 & 0.46 \\
& & & Thai
& $-3.7$ [$-3.9$, $-3.6$]
& $-14.8$ [$-15.1$, $-14.4$]
& $-2.58$ [$-2.65$, $-2.52$]
& 74 & 0.38 \\
& & & Thai (adjusted)
& $-3.5$ [$-3.6$, $-3.4$]
& $-14.0$ [$-14.3$, $-13.6$]
& $-2.45$ [$-2.51$, $-2.39$]
& 75 & 0.42 \\
\midrule
EN--JA & 11,217 & 4.9 & English
& $-4.1$ [$-4.3$, $-3.9$]
& $-12.7$ [$-13.3$, $-12.2$]
& $-2.24$ [$-2.35$, $-2.14$]
& 74 & 0.51 \\
& & & Japanese
& $-4.3$ [$-4.5$, $-4.1$]
& $-12.6$ [$-13.1$, $-12.0$]
& $-2.25$ [$-2.35$, $-2.15$]
& 69 & 0.46 \\
\midrule
EN--KO & 6,875 & 4.4 & English
& $-4.3$ [$-4.5$, $-4.1$]
& $-12.1$ [$-12.7$, $-11.5$]
& $-2.13$ [$-2.23$, $-2.02$]
& 75 & 0.51 \\
& & & Korean
& $-4.2$ [$-4.5$, $-3.9$]
& $-11.9$ [$-12.6$, $-11.2$]
& $-2.08$ [$-2.21$, $-1.95$]
& 73 & 0.51 \\
\bottomrule
\end{tabular*}
\caption{
Pitch trends estimated from utterances with
at least three aligned clusters.
For each recording, we fit a straight line
through the cluster mean pitches in their
original temporal order.
The per-cluster slope measures the fitted
pitch change between successive clusters,
expressed as a percentage of the recording's
utterance-mean $F_0$.
Because cluster counts differ across languages,
these slopes should not be compared directly.
The first-to-last columns instead summarize
the fitted change across the analyzed utterance,
from its first to its last cluster, providing
an utterance-wide measure for comparison.
This change is reported both as a percentage
of mean $F_0$ and in semitones, a logarithmic
pitch scale.
Negative values indicate falling pitch.
``Falling'' gives the percentage of utterances
with a negative fitted slope, while $R^2$
describes the proportion of variation in
cluster mean pitch explained by the line.
Brackets give episode-clustered 95\% confidence
intervals for the mean slopes and changes.
Adjusted rows use tone-adjusted pitch;
French uses the current processing settings.}
\label{tab:pitch_trends}
\end{table*}

\subsection{Detrending Sensitivity}
\label{app:detrending_sensitivity}

\begin{figure*}[t]
\centering
\includegraphics[width=\textwidth]
    {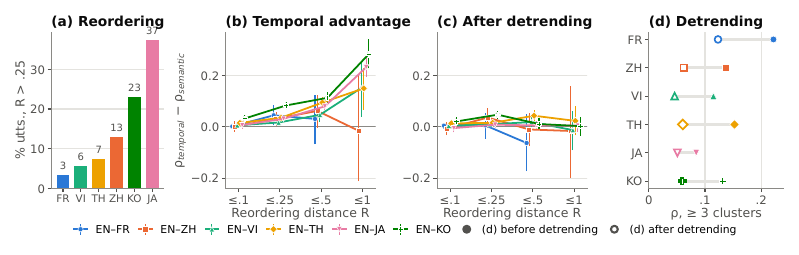}
\caption{
Cluster reordering and detrending sensitivity.
(a) Proportion of utterances with
normalized reordering distance $R>0.25$.
(b) Temporal-minus-semantic pitch correlation
differences across reordering-distance bins.
(c) The same comparison after removing
linear trends fitted to cluster means
in each recording's temporal order.
(d) Mean semantic-pairing correlations
before and after detrending.
Reordering comparisons use utterances
with at least two clusters; panel d
uses utterances with at least three.
French uses the current processing settings.}
\label{fig:reordering_detrending}
\end{figure*}

Figure~\ref{fig:reordering_detrending}
shows how the temporal advantage varies
with reordering distance before and
after detrending.
On each side, we remove a linear
trend fitted to cluster means over
the clusters' positions in that recording's
own temporal order.
We then repeat both pairing conditions.

\begin{figure}[t]
\centering
\includegraphics[width=\columnwidth]
    {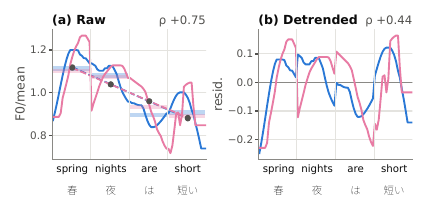}
\caption{
Illustration of broad pitch trends and
residual correspondence.
(a) Raw English and Japanese contours
under semantic pairing, with fitted
temporal trends shown by dashed lines.
(b) Paired contours after removing
each recording's temporal trend.
Blue denotes English and pink Japanese.}
\label{fig:detrending_example}
\end{figure}

Table~\ref{tab:detrended_correlations} reports
semantic-pairing correlations before and after
trend removal, together with the raw first-to-last
pitch changes.
Mean correlations decrease for every pair:
French falls from 0.222 to 0.123, while the
Asian-language pairs fall from 0.084--0.153
to 0.046--0.062.
Japanese and Korean retain means of 0.051 and
0.060, respectively, within the tonal-language
range; French remains higher.
Figure~\ref{fig:detrending_example} illustrates
a shared declining trend and positive residual
correspondence in an English--Japanese example.

The temporal advantage is substantially
smaller after detrending, particularly
for Japanese and Korean.
Their mean paired advantages decrease
from 0.090 to 0.005 and from 0.116
to 0.023, respectively
(Table~\ref{tab:pairing_detrending}).
The Japanese interval includes zero
after detrending, while positive
advantages with intervals excluding
zero remain for Vietnamese, Thai,
and Korean.
Removing broad linear trends therefore
reduces sensitivity to cluster ordering,
although the pairing conditions do
not become identical in every language.

The results support the interpretation
that semantic realignment can disrupt
shared sentence-level pitch trends.
These trends are meaningful prosodic
structure, and detrending also removes
genuine position-linked effects such
as final lowering.
It is therefore a sensitivity analysis,
rather than a complete separation of
sentence-level and local prosody.
Residual semantic-pairing means use
utterances with at least three clusters:
with two clusters, the fitted trend
removes all between-cluster mean differences.
The detrended Asian-language means lie
within a similar range, while the
French mean remains higher.

\section{Part-of-Speech Contributions}
\label{app:pos_results}

\begin{figure*}[!t]
\centering
\includegraphics[width=\textwidth]{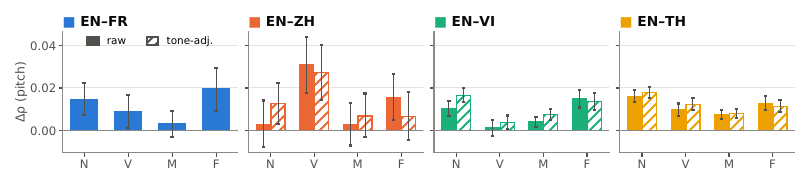}
\caption{
POS contributions to pitch correspondence
before and after tone adjustment.
Each contribution is the full correlation
minus the correlation after removing
the group's frames from both contours.
N, V, M, and F denote nouns, verbs,
modifiers, and function words, respectively.
Error bars indicate 95\% confidence
intervals for individual contributions.
French uses the current processing settings.}
\label{fig:pos_contributions}
\end{figure*}

We examine how word classes contribute to
word-aligned pitch correspondence.
Alignment clusters are assigned to nouns, verbs,
modifiers, or function words using their dominant
English-side POS tags from spaCy.
For each group, we remove its frames from both
contours and subtract the resulting correlation
from the full correlation.
A positive contribution indicates that removing
the group reduces correspondence.
Groups are evaluated separately, so their
contributions are not additive.
We repeat the analysis before and after
lexical-tone adjustment.
The French comparison uses a separate
10,000-utterance reference sample processed
with the same settings.

Figure~\ref{fig:pos_contributions} summarizes
the results, and Table~\ref{tab:pos_contributions}
reports the corresponding values.
Before adjustment, verbs contribute most
in Chinese, function words in Vietnamese,
and nouns in Thai.
After adjustment, noun contributions increase
from 0.003 to 0.013 in Chinese,
from 0.010 to 0.017 in Vietnamese,
and from 0.016 to 0.018 in Thai.
All three adjusted noun contributions have
95\% confidence intervals excluding zero
and are numerically close to or above
the French value of 0.015.

Modifier contributions also increase in all
three tonal languages.
Function-word contributions decrease most
in Chinese, from 0.016 to 0.007,
with the adjusted interval including zero.
Verb contributions decrease in Chinese
and increase in Vietnamese and Thai.
These changes describe how tone adjustment
shifts the distribution of POS contributions;
the reported intervals assess individual
contributions, not the significance of
raw-to-adjusted changes.


\begin{table}[H]
\centering
\small
\setlength{\tabcolsep}{3pt}
\begin{tabular*}{\columnwidth}{
@{\extracolsep{\fill}}lcccc@{}}
\toprule
\textbf{Pair / condition}
& \textbf{Nouns}
& \textbf{Verbs}
& \textbf{Mod.}
& \textbf{Func.} \\
\midrule
EN--FR
& $+.015^{*}$ & $+.009^{*}$
& $+.003$ & $+.020^{*}$ \\
\midrule
EN--ZH
& $+.003$ & $+.031^{*}$
& $+.003$ & $+.016^{*}$ \\
\quad Adjusted
& $+.013^{*}$ & $+.027^{*}$
& $+.007$ & $+.007$ \\
\midrule
EN--VI
& $+.010^{*}$ & $+.001$
& $+.004^{*}$ & $+.015^{*}$ \\
\quad Adjusted
& $+.017^{*}$ & $+.004^{*}$
& $+.008^{*}$ & $+.014^{*}$ \\
\midrule
EN--TH
& $+.016^{*}$ & $+.010^{*}$
& $+.007^{*}$ & $+.013^{*}$ \\
\quad Adjusted
& $+.018^{*}$ & $+.012^{*}$
& $+.008^{*}$ & $+.011^{*}$ \\
\bottomrule
\end{tabular*}
\caption{
Leave-one-group-out pitch contributions.
Unmarked rows use raw pitch; adjusted rows
use tone-adjusted pitch.
An asterisk indicates that the individual
contribution's 95\% confidence interval
excludes zero.
It does not test the change between
raw and adjusted contributions.
Mod.\ and Func.\ denote modifiers
and function words.
French uses the current processing settings
on the separate POS reference sample.}
\label{tab:pos_contributions}
\end{table}

\section{Additional Correlation Analyses}
\label{app:additional_correlations}

\subsection{Utterance-Level Distributions}
\label{app:correlation_distributions}

\begin{figure*}[t]
\centering
\includegraphics[width=\textwidth]
    {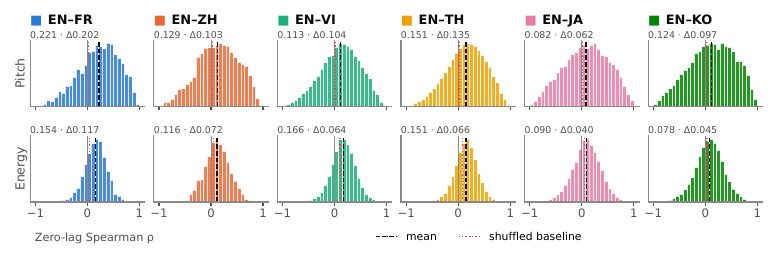}
\caption{
Distributions of utterance-level raw
pitch and energy correlations.
Black dashed lines mark mean correlations,
and red dotted lines mark unit-shuffled
baseline means.
Each annotation gives the mean correlation
and its difference above baseline.
These differences use unrounded estimates.
French uses the current processing settings.}
\label{fig:correlation_distributions}
\end{figure*}

Figure~\ref{fig:correlation_distributions}
shows the distributions of raw pitch
and energy correlations across utterances.
The distributions contain both positive
and negative scores, despite their
positive means.
The vertical lines distinguish each
pair's mean from its unit-shuffled
baseline.
The annotations report the mean and
the mean difference above baseline;
the latter is distinct from the
tone-adjustment change used in the
main results.

\subsection{Matched Tonal-Pair Comparisons}
\label{app:tonal_pair_figure}

\begin{figure*}[t]
\centering
\includegraphics[width=\textwidth]
    {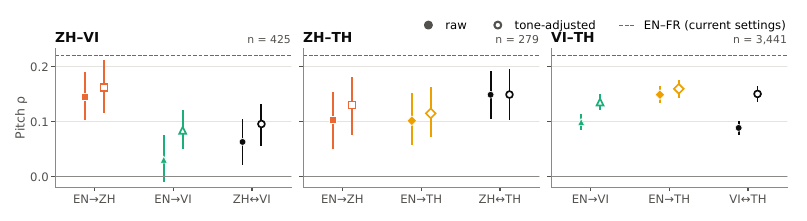}
\caption{
Raw and tone-adjusted pitch correlations
for matched tonal-pair cohorts and
their corresponding English comparisons.
Filled markers indicate raw pitch;
open markers indicate adjusted pitch.
Both recordings are adjusted in the
direct tonal-pair comparisons, while
only the tonal recording is adjusted
in the English comparisons.
Each language pair is aligned separately.
Dashed lines mark the current French
reference, which is outside these cohorts.}
\label{fig:tonal_pairs}
\end{figure*}

Figure~\ref{fig:tonal_pairs} shows
the direct tonal-pair correlations
alongside the corresponding English
comparisons on the matched cohorts.
English comparisons provide context
for the direct-pair analysis; each
language pair is aligned separately.
The correlations therefore describe
pair-specific contour constructions,
rather than a single alignment shared
by all three recordings.

\begin{table}[H]
\centering
\small
\setlength{\tabcolsep}{3pt}
\begin{tabular*}{\columnwidth}{
@{\extracolsep{\fill}}lrrrr@{}}
\toprule
& \multicolumn{2}{c}{
\shortstack{First-to-last\\pitch change (\%)}}
& \multicolumn{2}{c}{
\shortstack{Pitch correlation\\(mean $\rho$)}} \\
\cmidrule(lr){2-3}
\cmidrule(lr){4-5}
\textbf{Pair}
& \textbf{EN}
& \textbf{Other}
& \textbf{Before}
& \textbf{After} \\
\midrule
EN--FR & $-14.5$ & $-11.8$ & 0.222 & 0.123 \\
EN--ZH & $-12.3$ & $-10.1$ & 0.136 & 0.062 \\
EN--VI & $-14.1$ & $-6.8$  & 0.115 & 0.046 \\
EN--TH & $-13.8$ & $-14.8$ & 0.153 & 0.060 \\
EN--JA & $-12.7$ & $-12.6$ & 0.084 & 0.051 \\
EN--KO & $-12.1$ & $-11.9$ & 0.131 & 0.060 \\
\bottomrule
\end{tabular*}
\caption{
Within-utterance pitch decline and correspondence
before and after detrending.
Pitch change is the mean fitted first-to-last
change as a percentage of utterance-mean $F_0$,
reported separately for English (EN) and the
other language; negative values indicate decline.
Each recording's trend is fitted through its
cluster mean pitches in its own temporal order.
The final columns report mean semantic-pairing
correlations before and after removing these trends.
Each pair uses its own utterances with at least
three aligned clusters, with the same utterances
scored before and after detrending.}
\label{tab:detrended_correlations}
\end{table}

\begin{table}[H]
\centering
\small
\setlength{\tabcolsep}{3pt}
\begin{tabular*}{\columnwidth}{
@{\extracolsep{\fill}}lrrrr@{}}
\toprule
& \multicolumn{2}{c}{\textbf{Duration}}
& \multicolumn{2}{c}{\textbf{Phone count}} \\
\cmidrule(lr){2-3}
\cmidrule(lr){4-5}
\textbf{Pair}
& \textbf{$n$} & \textbf{$\rho$}
& \textbf{$n$} & \textbf{$\rho$} \\
\midrule
EN--FR & 1,931  & 0.938 & 1,931  & 0.905 \\
EN--ZH & 1,318  & 0.758 & 1,318  & 0.730 \\
EN--VI & 17,452 & 0.916 & 17,452 & 0.888 \\
EN--TH & 21,183 & 0.847 & 21,120 & 0.837 \\
\bottomrule
\end{tabular*}
\caption{
Cross-utterance Spearman correlations
in duration and canonical phoneme count
at the primary SONAR threshold of 0.45.
These are correlations across utterances,
rather than means of within-utterance
contour correlations.
French uses the current processing settings.}
\label{tab:temporal_correlations}
\end{table}

\subsection{Duration and Phoneme Counts}
\label{app:temporal_results}

Table~\ref{tab:temporal_correlations}
reports cross-utterance correlations
in duration and canonical phoneme count.
Duration is measured from timed transcript
extent, and phoneme counts come from
grapheme-to-phoneme conversion of the
recognized text.
We retain pairs with equal positive
sentence counts for these comparisons.
The Thai phoneme-count cohort is smaller
because some pairs lack valid counts.

Duration correlations range from 0.758
to 0.916 for the English--tonal pairs,
and phoneme-count correlations from
0.730 to 0.888.
These results indicate correspondence
in coarse temporal and transcript-derived
quantities alongside the lower contour
correlations.
Canonical counts describe expected
pronunciations of the recognized text,
rather than acoustically measured
realized phonemes.

\section{Filtering and Feature Sensitivity}
\label{app:robustness}

\subsection{Semantic Filtering Sensitivity}
\label{app:sonar_sensitivity}

We repeat the pitch analysis at SONAR
0.63 rather than the primary threshold
of 0.45.
This is the semantic threshold used
by \citet{xie2026prosodylosttranslationfinegrained}.
Raw and adjusted means for the tonal
languages remain below the current
French reference at this threshold
(Table~\ref{tab:sonar_sensitivity}).
The descriptive difference therefore
persists under stricter semantic filtering.

\begin{table}[H]
\centering
\small
\setlength{\tabcolsep}{4pt}
\begin{tabular*}{\columnwidth}{
@{\extracolsep{\fill}}lcc@{}}
\toprule
\textbf{Pair}
& \textbf{Raw $\rho$}
& \textbf{Adjusted $\rho$} \\
\midrule
EN--FR & 0.225 & --- \\
\midrule
EN--ZH & 0.154 & 0.163 \\
EN--VI & 0.123 & 0.153 \\
EN--TH & 0.158 & 0.174 \\
\bottomrule
\end{tabular*}
\caption{
Mean pitch correlations at SONAR 0.63.
Raw and adjusted values use their
respective available scoring cohorts;
their differences are not paired
adjustment estimates.
French uses the current processing settings.}
\label{tab:sonar_sensitivity}
\end{table}

\subsection{Alignment Benchmarks}
\label{app:alignment_benchmarks}

Table~\ref{tab:alignment_benchmarks}
reports precision, recall, and alignment
error rate against human gold links.
The Japanese benchmark \citep{japan-gold} has lower recall
than the Chinese benchmark \citep{chinese-gold}, making
missed links a particular concern
for the Japanese contour analysis.

\begin{table}[H]
\centering
\small
\setlength{\tabcolsep}{4pt}
\begin{tabular*}{\columnwidth}{
@{\extracolsep{\fill}}lrccc@{}}
\toprule
\textbf{Language}
& \textbf{$n$}
& \textbf{Precision}
& \textbf{Recall}
& \textbf{AER} \\
\midrule
Chinese  & 900   & 80.47\% & 48.52\% & 39.42\% \\
Japanese & 1,235 & 70.95\% & 22.23\% & 66.15\% \\
\bottomrule
\end{tabular*}
\caption{
Alignment performance against human
gold links in text benchmarks.
AER denotes alignment error rate.
$n$ is the number of benchmark sentences.
The Chinese benchmark contains newswire
and the Japanese benchmark Wikipedia text.}
\label{tab:alignment_benchmarks}
\end{table}

These evaluations use text outside
the dubbing domain.
Benchmark sentence text is included
in unsupervised alignment training,
while gold links are withheld.
The scores characterize this transductive
benchmark setting and do not directly
measure alignment accuracy in the
analyzed recordings.

\subsection{Contour Resolution and Track Checks}
\label{app:feature_checks}

Varying contour resolution from 10
to 40 frames per aligned cluster
changes raw-pitch mean correlations
by at most approximately 0.0021
in the reported checks.
The Thai and Vietnamese checks use
samples of roughly 3,000 pairs.
Adjusted-pitch resolution checks show
a spread of at most 0.002.

A scan of 169,522 saved raw-pitch
tracks identifies five containing
negative interpolated pitch values.
Excluding the affected utterance pairs
changes the corresponding mean correlations
by less than 0.0002.
Checks on 300 utterances per language
also confirm that tone adjustment
preserves time coordinates and voicing,
and leaves frames outside labeled
syllables unchanged.

\subsection{Contour Construction Example}
\label{app:contour_example}

\begin{figure*}[t]
\centering
\includegraphics[width=\textwidth]
    {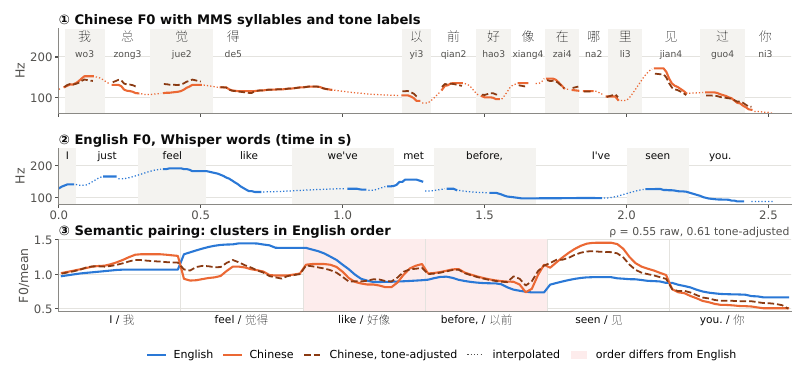}
\caption{
Chinese--English contour construction.
Top: Chinese pitch with MMS syllable
intervals and tone labels.
Middle: English pitch with timed
transcript units.
Bottom: normalized contours for selected
alignment clusters in English order,
including raw and adjusted Chinese pitch.
Shading marks clusters whose order
differs between recordings; dotted
segments indicate interpolated pitch.
Displayed links are automatically generated.}
\label{fig:contour_example}
\end{figure*}

Figure~\ref{fig:contour_example}
illustrates contour construction for
a Chinese--English utterance.
It shows the syllable intervals used
for tone adjustment, the timed English
transcript, and the selected alignment
clusters in English order.
The example illustrates the automatic
procedure rather than a manually
corrected alignment.

\subsection{Statistical Dependence}
\label{app:statistical_dependence}

Confidence intervals use episode-clustered
bootstrapping, retaining the utterances
within each sampled episode together.
Episodes can also share casts and
recording conditions, and English
segments can recur across language
comparisons.
These dependencies beyond the episode
are not fully represented by the
reported intervals or utterance-level
rank tests.

\section{Estimated Voice-Label Distribution}
\label{app:gender_distribution}

We use inaSpeechSegmenter \citep{gendertool} obtain automatic
male/female voice labels for both recordings
in each scored English--tonal pair and the
English--French reference.
Table~\ref{tab:gender_distribution} summarizes
the distributions.
These are acoustic classifications, not
self-reported gender identities.
Processing completed without errors;
percentages exclude recordings with no
detected speech (4--96 per language side).
Japanese and Korean have not yet been evaluated.

\begin{table}[H]
\centering
\small
\setlength{\tabcolsep}{3pt}
\begin{tabular*}{\columnwidth}{
@{\extracolsep{\fill}}lrrr@{}}
\toprule
\textbf{Pair}
& \textbf{English M/F}
& \textbf{Target M/F}
& \textbf{Pairs} \\
\midrule
EN--FR & 55.1 / 44.9 & 54.0 / 46.0 & 2,713 \\
EN--ZH & 62.7 / 37.3 & 61.9 / 38.1 & 1,838 \\
EN--VI & 57.2 / 42.8 & 56.2 / 43.8 & 22,332 \\
EN--TH & 57.0 / 43.0 & 55.6 / 44.4 & 31,261 \\
\bottomrule
\end{tabular*}
\caption{
Automatic male/female (M/F) voice-label
percentages for the scored utterance pairs.
Percentages exclude recordings with no
detected speech; pair counts include the
full scored cohorts.
French uses the current processing settings.}
\label{tab:gender_distribution}
\end{table}

English and target labels agree for
approximately 94--95\% of pairs.
Across the evaluated language sides,
median $F_0$ ranges from 113--137~Hz
for male-labeled recordings and
180--225~Hz for female-labeled recordings.
These summaries provide descriptive checks
on the automatic labels; they do not
establish classification accuracy or
control for speaker composition.